\documentclass[]{fairmeta}
\usepackage{graphicx}
\usepackage[utf8]{inputenc}
\usepackage[T1]{fontenc}
\usepackage{amsmath}
\usepackage{amssymb}
\usepackage{mathtools}
\usepackage{bm}
\usepackage{algorithm}
\usepackage{algpseudocode}
\usepackage{paralist}
\usepackage[most]{tcolorbox}
\usepackage{booktabs}
\usepackage{array}
\usepackage{tabularx}
\usepackage{ragged2e}
\usepackage{makecell}
\usepackage{multirow}
\usepackage{tikz}
\usepackage{pgfplots}
\usepackage{wrapfig}
\usetikzlibrary{pgfplots.groupplots, matrix}
\pgfplotsset{compat=1.18}
\usepackage{enumitem}
\usepackage{comment}
\usepackage{graphicx}
\usepackage{multirow}
\usepackage{stackengine}
\usepackage{colortbl} 
\usepackage{anyfontsize}
\usepackage{CJK}
\usepackage{subfig}
\usepackage{placeins}

\definecolor{purple}{HTML}{c994c7}
\definecolor{navyblue}{RGB}{30,130,255}
\definecolor{citecolor}{RGB}{30,130,255}
\definecolor{lightgray}{gray}{0.9}
\definecolor{blanchedalmond}{rgb}{1.0, 0.92, 0.8}
\definecolor{cerise}{rgb}{0.871, 0.192, 0.388}

\definecolor{TaskBG}{HTML}{EFE6FF}        
\definecolor{StateBG}{HTML}{F5F5F7}       
\definecolor{ExpertBG}{HTML}{EAF7EA}      
\definecolor{IWMBG}{HTML}{FDECF3}         
\definecolor{SRBG}{HTML}{E6F2FF}          

\usepackage[utf8]{inputenc}
\definecolor{lastauthor}{RGB}{143, 68, 115}

\title{\includegraphics[height=1.2em]{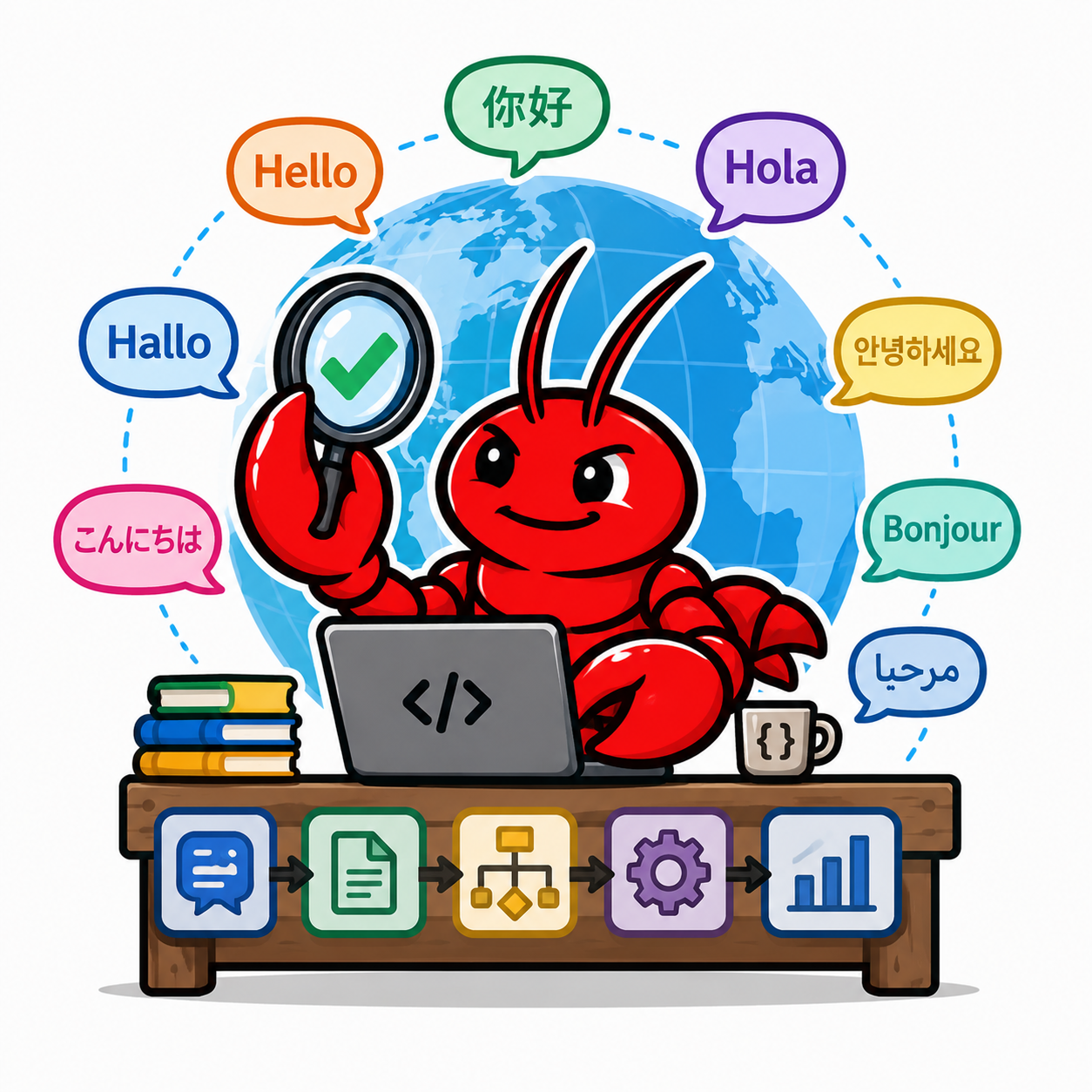}\hspace{0.3em}PolyWorkBench: Benchmarking LLM Agents for Cross-Lingual Long-Horizon Workflows}

\author[1,2]{Hongliang Li*}
\author[2]{Yijin Liu}
\author[1,2]{Zhiwei Zhang*}
\author[2]{Zihe Liu}
\author[1]{Xinyue Lou}
\author[1]{Jinan Xu}
\author[2]{Fandong Meng}
\author[1]{Kaiyu Huang}

\affiliation[1]{Beijing Jiaotong University}
\affiliation[2]{Weixin AI, Tencent Inc}
\contribution[*]{Work done during an internship at Weixin AI, Tencent Inc.}

\abstract{
While Large Language Model (LLM) agents excel at monolingual long-horizon planning and tool use, enterprise workflows inherently require processing multilingual resources across extended trajectories. The interaction between multilinguality and long-horizon execution, however, remains underexplored. We introduce \textbf{PolyWorkBench}, a benchmark designed to evaluate LLM agents on multilingual, long-horizon workplace workflows. PolyWorkBench features 67 tasks across five core domains: commerce, knowledge work, legal analysis, localization, and manufacturing. Tasks are authored by the paper's authors from real-world data seeds and independently verified through a second-author audit. Agents must integrate heterogeneous multilingual inputs, execute iterative tool-use trajectories, and produce structured domain artifacts. To rigorously assess performance, we adopt \textit{Grade}, a task-specific structural scoring rubric, as our primary ranking metric, and complement it with \textit{Pytest} for executable state verification and \textit{LLM-as-Judge} for semantic quality diagnostics. Benchmark evaluations reveal that agent performance varies substantially across languages and drops sharply on the harder cross-lingual tasks, and our analysis shows that multilingual execution exposes systematic failure modes across planning, tool interaction, and decision-making in long-horizon agents.
}

\date{July 7, 2026}
\correspondence{Hongliang Li at \email{superhanslerli@gmail.com}, Kaiyu Huang at \email{kyhuang@bjtu.edu.cn}}

\begin{document}

\maketitle

\section{Introduction}

\begin{figure*}[t]
    \centering
    \includegraphics[width=\textwidth]{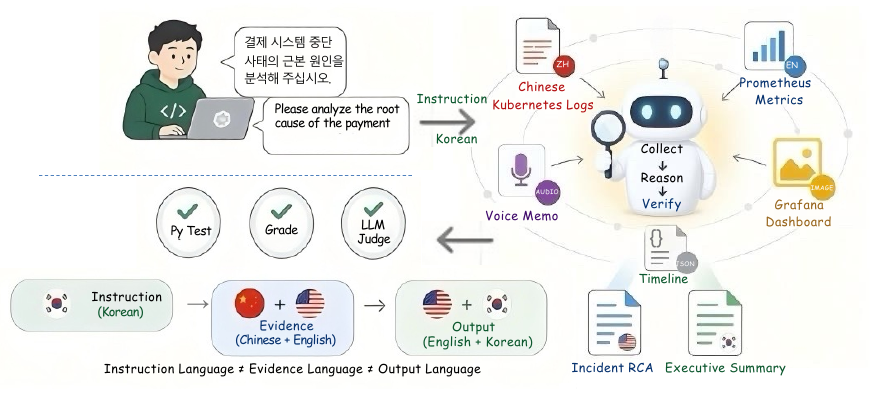}
    \caption{
Overview of a representative task (MFG-00). A single task incorporates all four benchmark axes: cross-lingual trajectory decomposition, long-horizon tool use, cross-source fusion, and multi-faceted ground-truth scoring.}
    \label{fig:intro}
\end{figure*}

Recent advances in large language model (LLM) agents have enabled substantial progress on long-horizon tasks that require planning, tool use, and environment interaction~\cite{xu2025llmsur,yehudai2026survey}. Existing workflow-oriented benchmarks (e.g., SWE-bench~\cite{jimenez2024swe}, WebArena~\cite{zhou2024webarena}, OSWorld~\cite{xie2024osworld}, OdysseyBench~\cite{wang2025odysseybench}, CoffeeBench~\cite{sugiura2026coffeebench}) have been developed to probe sequential decision-making over extended horizons. However, these long-horizon benchmarks predominantly assume a monolingual (typically English) environment, neglecting the complex cross-lingual interactions ubiquitous in real-world workplaces. Conversely, traditional multilingual benchmarks (e.g., XGLUE~\cite{liang2020xglue}, M-MMLU~\cite{lai2023okapi}, MGSM~\cite{shi2022language}) focus primarily on single-turn or short-horizon tasks like cross-lingual QA or reasoning, failing to capture the dynamic multilingual execution required in autonomous agent workflows. This reveals a critical gap: evaluating how agents execute long-horizon workflows across dynamic, multilingual environments.

These two research directions treat \emph{multilinguality} and \emph{agentic reasoning} as orthogonal dimensions. Agent benchmarks are almost exclusively monolingual: task instructions, intermediate reasoning, tool interactions, and final outputs are typically English, and multilinguality, when present at all, is reduced to surface-level translation. Multilingual benchmarks, conversely, are confined to isolated static predictions, where each instance is independent and does not involve multi-step interaction or tool-based execution. As a result, one important question is left unanswered: how language variation affects the stability of reasoning and decision-making in long-horizon agent workflows.

Enterprise workflows routinely span multiple languages: users may issue instructions in one language, retrieve evidence in another, interact with tools expecting a third, and finally produce multilingual deliverables. Such workflows require language variation to persist throughout planning, tool use, and artifact generation, rather than appearing only at the input or output stages. However, existing benchmarks do not capture this execution setting~\cite{jimenez2024swe, zhou2024webarena, li2026claw}.  Figure~\ref{fig:intro} illustrates a representative workflow requiring cross-lingual planning, tool interaction, and multilingual artifact generation.

To study this setting, we introduce PolyWorkBench, a benchmark of realistic multilingual workplace workflows together with a hybrid evaluation protocol for assessing long-horizon agents. Experiments show that agents score substantially lower in multilingual long-horizon execution than under monolingual conditions, and that multilingual execution exposes failure modes which remain hidden under monolingual evaluation. 
Our contributions are three-fold:

\begin{itemize}

\item We identify multilingual long-horizon agent evaluation as a previously overlooked benchmark setting, in which language variation permeates the entire execution workflow rather than appearing only in task inputs or outputs.

\item We introduce PolyWorkBench, a benchmark of 67 realistic workplace workflows spanning five domains and ten languages, together with a hybrid evaluation protocol combining structural grading, executable verification, and semantic assessment.

\item We conduct the first comprehensive study of multilingual long-horizon agents. Multilingual execution exposes systematic failure modes across planning, tool interaction, and workflow completion. These failure modes remain hidden under monolingual evaluation.

\end{itemize}

\section{Related Work}
\label{sec:related-work}

\textbf{Agent and Multilingual Benchmarks.} The emergence of large language models has driven increasingly realistic benchmarks for evaluating autonomous agents. Early efforts such as AgentBench~\cite{liu2024agentbench} established unified evaluation across interactive environments. Subsequent benchmarks expanded to web navigation (WebArena~\cite{zhou2024webarena}), open-world assistant tasks (GAIA~\cite{mialon2024gaia}), and desktop operation (OSWorld~\cite{xie2024osworld}), while domain-specific benchmarks such as SWE-bench~\cite{jimenez2024swe}, MLE-bench~\cite{chan2025mle}, and $\tau$-bench~\cite{yao2024tau} evaluate software engineering and tool-use scenarios. Recent workflow-oriented benchmarks including Claw-Eval~\cite{ye2026claw}, Claw-Eval-Live~\cite{li2026claw}, WildClawBench~\cite{ding2026wildclawbench}, and ClawMark~\cite{meng2026clawmark} emphasize persistent environments and evolving workflows. In parallel, multilingual benchmarks such as FLORES~\cite{goyal2022flores}, MGSM~\cite{shi2022language}, M-MMLU~\cite{lai2023okapi}, and MMMLU~\cite{hendrycks2020measuring} assess multilingual understanding and reasoning, and recent efforts including MAPS~\cite{hofman2026maps} and MIRAGE-Bench~\cite{thakur2025mirage} extend to multilingual multi-step tasks. Even these multilingual multi-step benchmarks, however, keep language fixed within an instance: MAPS varies the language across problems but each problem is solved in a single language, and MIRAGE-Bench evaluates multilingual retrieval-augmented generation where the language of a query and its evidence is held constant rather than shifting across the trajectory. Neither exercises the setting where the instruction language, the resource languages, and the output language differ within one task and language transitions occur during planning, tool interaction, and artifact generation, which is precisely the trajectory-level multilinguality PolyWorkBench targets. Task-oriented benchmarks like SpreadsheetBench~\cite{ma2024spreadsheetbench}, OfficeBench~\cite{wang2024officebench}, and RAGBench~\cite{friel2024ragbench} probe structured-document and retrieval-intensive workflows. Compared with these lines of work, PolyWorkBench integrates multilingual understanding, heterogeneous document processing, long-horizon reasoning, and structured output generation into unified end-to-end enterprise tasks, treating language variation as a trajectory-level property rather than an input attribute.

\noindent \textbf{Benchmark Evaluation.} As benchmarks become more open-ended, evaluation has shifted from rule-based metrics toward model-based assessment. Traditional metrics such as exact match are often insufficient for generative and multi-turn tasks. LLM-as-a-judge has therefore emerged as a scalable paradigm: representative systems such as MT-Bench and Chatbot Arena~\cite{zheng2023judging} show that strong LLM judges achieve high agreement with human judgments on open-ended outputs, and AlpacaEval~\cite{alpaca_eval} enables efficient pairwise ranking of instruction following. However, prior work has identified limitations including position bias, verbosity bias, and prompt sensitivity~\cite{zheng2023judging}, raising concerns about robustness in complex or long-horizon settings. Reliable evaluation of multilingual multi-step agent workflows therefore remains an open challenge, motivating the deterministic-plus-semantic evaluation framework we adopt in PolyWorkBench.

\section{PolyWorkBench}

\subsection{Task Design}

\begin{figure*}[t]
    \centering
    \includegraphics[width=\textwidth]{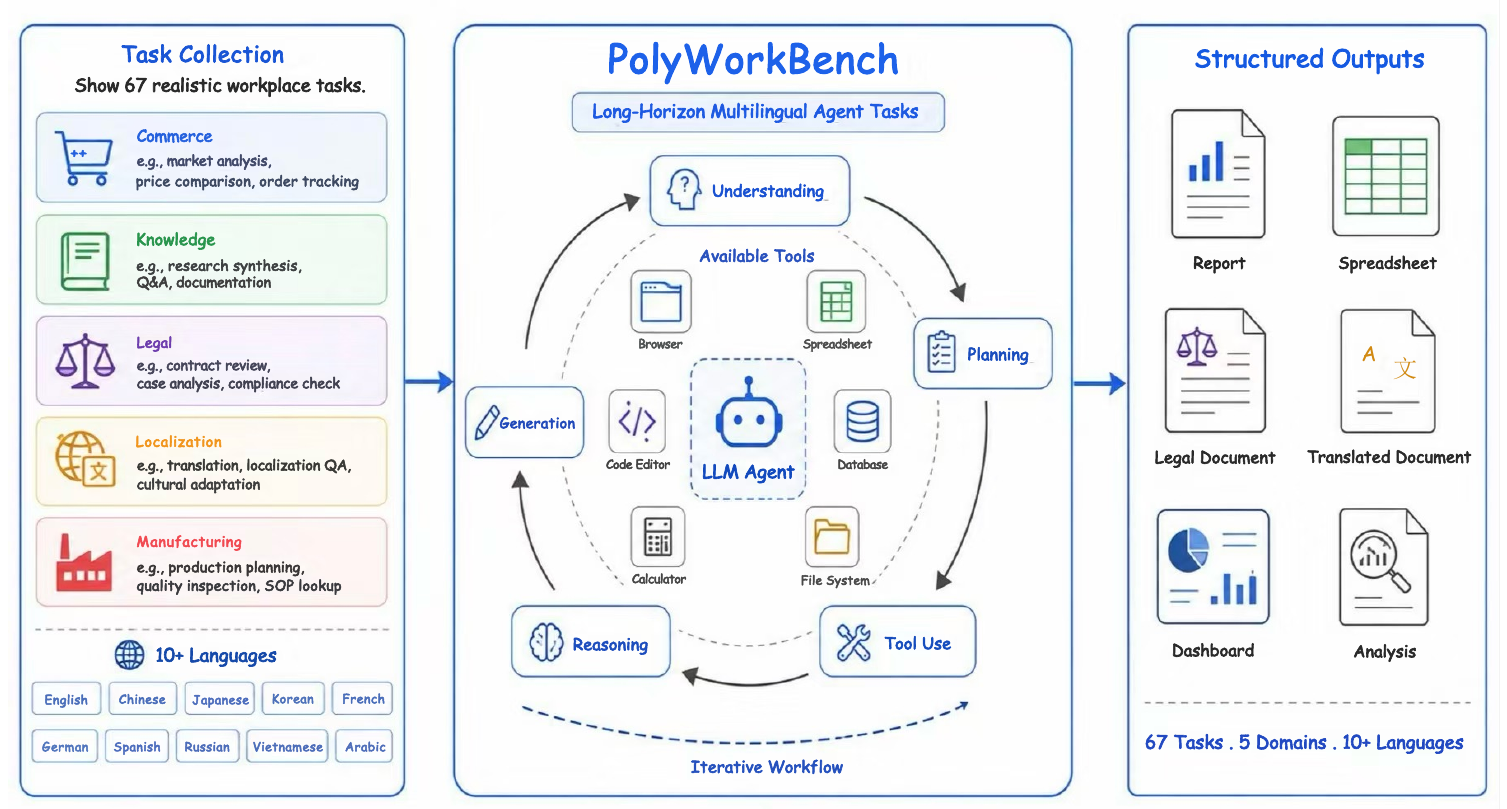}
    \caption{
    Overview of PolyWorkBench. The benchmark contains 67 multilingual long-horizon workplace tasks spanning five representative domains. Given heterogeneous multilingual inputs, an LLM agent performs iterative reasoning and tool use to generate structured outputs for downstream evaluation.
    }
    \label{fig:overview}
\end{figure*}

Following the system-oriented formulation of long-horizon agents~\cite{dong2026towards}, we consider an evaluated agent as the coupled system

\begin{equation}
\text{Agent}=\pi_{\theta}\oplus H,
\label{eq:agent}
\end{equation}

where $\pi_{\theta}$ denotes the foundation-model policy and $H$ the external runtime harness (planning, memory, tool invocation, orchestration, verification). Under this formulation, long-horizon capability is a property of the coupled system rather than the foundation model alone.

Accordingly, PolyWorkBench evaluates multilingual long-horizon agents through complete workplace workflows rather than isolated reasoning problems. We model each benchmark instance as a workflow $\mathcal{W}=(\mathcal{T}, \mathcal{R}, \tau, \mathcal{E})$, where $\mathcal{T}$ is the task specification, $\mathcal{R}$ heterogeneous multilingual resources, $\tau=(s_1,\ldots,s_n)$ the execution trajectory with each state $s_i=(p_i,a_i,o_i)$ combining planning, action, and observation, and $\mathcal{E}$ the evaluation protocol. Benchmark evaluation is then $\mathcal{P}=\Phi(\pi_{\theta}\oplus H, \mathcal{W})$, where $\Phi(\cdot)$ is the unified evaluation framework of Section~\ref{sec:evaluation}. Full trajectory-level formalism is deferred to App.~\ref{app:formalism}.

Unlike conventional multilingual benchmarks that evaluate language understanding at isolated prediction points, PolyWorkBench evaluates whether the coupled agent system can complete an entire multilingual execution trajectory while maintaining semantic consistency across planning, tool interaction, information integration, and structured artifact generation. Each task represents a complete business workflow in which an agent receives practical instructions together with supporting materials (documents, spreadsheets, logs, contracts, or multilingual resources) and must produce deliverables satisfying predefined requirements. Tasks require long-horizon reasoning across multiple intermediate steps: agents must retrieve information from heterogeneous resources, perform multilingual comprehension and translation, invoke external tools, integrate intermediate evidence, and generate structured outputs. Language variation therefore becomes an intrinsic property of the execution trajectory rather than a superficial attribute of task inputs or outputs.

PolyWorkBench covers five representative workplace domains: \textbf{Commerce} (cross-border operations, pricing, logistics, taxation, marketplace management), \textbf{Knowledge} (information synthesis, technical reporting, patent analysis, multilingual fact verification), \textbf{Legal} (compliance analysis, contract review, regulatory comparison, legal drafting), \textbf{Localization} (multilingual adaptation of software, documentation, subtitles, marketing materials), and \textbf{Manufacturing} (quality management, production reporting, maintenance, safety auditing, supply-chain analysis). Collectively these domains provide broad coverage of realistic multilingual workplace workflows requiring long-horizon reasoning, external tool use, and structured artifact generation.

\subsection{Data Overview}
\label{sec:data-overview}  

PolyWorkBench contains 67 manually designed evaluation tasks spanning five representative workplace domains and ten languages,
including English,
Chinese,
Japanese,
Korean,
French,
German,
Spanish,
Russian,
Vietnamese,
and Arabic.
Rather than evaluating monolingual execution,
many benchmark instances require reasoning across multiple languages within a single workflow,
reflecting the multilingual nature of modern enterprise environments.
Language coverage is intentionally uneven and follows the distribution of realistic workplace artefacts rather than a uniform sampling target,
so several languages (most notably Arabic, with a single task) are represented only sparsely.
We therefore treat PolyWorkBench as a probe of multilingual long-horizon execution on a focused set of high-frequency enterprise languages,
and we do not claim broad language-universal coverage.
Per-language counts are reported in App.~\ref{app:formalism} and sparsely covered languages are excluded from any claim that requires more than one task.

Despite their domain diversity, all benchmark instances follow the workflow abstraction $\mathcal{W}=(\mathcal{T},\mathcal{R},\tau,\mathcal{E})$ introduced in the previous subsection (see App.~\ref{app:formalism} for full details). Specifically, each instance consists of four core components: (1)~a task specification $\mathcal{T}$ describing the business objective; (2)~heterogeneous multilingual resources $\mathcal{R}$ (e.g., documents, spreadsheets, contracts, databases, and logs); (3)~executable evaluation scripts $\mathcal{E}$ for deterministic verification; and (4)~structured reference specifications $\tau$ defining task-specific evaluation criteria.

Instead of relying on a single reference answer, PolyWorkBench adopts task-specific evaluation specifications that jointly capture structural correctness, functional validity, and semantic quality, accommodating the open-ended nature of workplace workflows where multiple execution trajectories may produce valid solutions. Instance diversity arises from different execution trajectories, and reflects the number of reasoning stages, the diversity of multilingual resources, the extent of tool interaction, and the dependencies among intermediate artifacts. So PolyWorkBench evaluates agents under substantially different execution conditions while maintaining a unified formulation. Multilinguality is embedded throughout execution rather than treated as an isolated evaluation dimension: language transitions may occur during planning, document understanding, tool invocation, information integration, and artifact generation, making multilingual consistency a trajectory-level property. The benchmark is designed to remain extensible: new tasks can be added without modifying the workflow abstraction or the evaluation framework.

\paragraph{Construction and quality control.}
Each task is authored by the paper's authors and follows a three-stage pipeline. First, we collect a real-world data seed for the target domain, either a synthetically generated artefact or a document derived from publicly redistributable corpora, and use it to fix the business scenario, the required deliverable, and the languages of the instruction, the supporting resources, and the expected output. Second, an author drafts the task specification, assembles the multilingual resources, and writes the reference specification as a set of weighted structural sub-scores together with an executable test suite where deterministic verification applies. Third, a second author independently completes the task and audits the specification, checking that the instruction is unambiguous, that every sub-score is objectively verifiable from the on-disk artefact, and that at least one valid solution trajectory exists. Tasks that fail this audit are revised until both authors agree. All supporting resources use fictional names, addresses, and account numbers, and no task contains personally identifying information. Full provenance, licensing, and per-task statistics are reported in App.~\ref{app:formalism}.


\subsection{Evaluation Framework}

PolyWorkBench adopts a multi-dimensional evaluation framework anchored by \textbf{Grade} ($S_{\text{grade}}$) as the primary metric, complemented by two diagnostic signals: \textbf{Pytest} ($S_{\text{pytest}}$) and \textbf{LLM-as-Judge} ($S_{\text{judge}}$).

\textbf{Grade ($S_{\text{grade}}$): Primary Metric.} 
To ensure reproducible and fine-grained evaluation, each task $T_i$ is assigned a set of $K$ deterministic, audit-verified rubrics $\{r_k\}_{k=1}^K$ with non-negative weights $w_k$ ($\sum w_k = 1$). $S_{\text{grade}}$ measures structural task completion from on-disk artifacts via deterministic scripts:
\begin{equation}
S_{\text{grade}}(T_i) = \sum_{k=1}^{K} w_k \cdot \mathbb{I}(r_k(\mathcal{A}_i) = \text{Pass})
\end{equation}
where $\mathcal{A}_i$ is the generated artifact and $\mathbb{I}(\cdot)$ is an indicator function checking objective criteria (e.g., schema, matched values, source lines).

\textbf{Pytest ($S_{\text{pytest}}$): Executable Verification.} 
For tasks with executable outputs, automated test suites execute unit/integration tests to verify deterministic correctness:
\begin{equation}
S_{\text{pytest}}(T_i) = \frac{1}{|M_i|} \sum_{m \in M_i} \mathbb{I}(\text{Exec}(m, \mathcal{A}_i) = \text{Success})
\end{equation}
where $M_i$ is the set of test cases for task $T_i$.

\textbf{LLM-as-Judge ($S_{\text{judge}}$): Semantic Assessment.} 
To capture unstructured semantic quality (e.g., coherence, faithfulness) beyond rule-based checks, a standardized judge prompt $P_{\text{judge}}$ rates the output on a discrete scale:
\begin{equation}
S_{\text{judge}}(T_i) = \text{LLM}(P_{\text{judge}}, T_i, \mathcal{A}_i) \in [0, 1]
\end{equation}

\paragraph{Metric Disaggregation.} We intentionally avoid aggregating these metrics into a single composite score. $S_{\text{pytest}}$ acts as a strict structural subset of $S_{\text{grade}}$, while $S_{\text{judge}}$ exhibits bimodal behavior that could distort linear model rankings. Full rubric specifications are in App.~\ref{app:eval-protocol}.

\begin{table*}[t]
\centering
\small
\caption{Overall performance on PolyWorkBench pivoted as model$\times$harness. Each cell shows Pass@1$^{n_{\text{runs}}}$ (Pass@3), where Pass@1 is the best-run mean per-task Grade and Pass@3 is the per-task best-of-three Grade averaged over the top three completed runs. Best across the leaderboard in \textbf{bold}, second-best \underline{underlined}. Rows are ordered by peak Pass@1 across harnesses.}
\label{tab:main-results}
\vspace{2pt}
\begin{tabular}{lcccc}
\toprule
Model & ClaudeCode & OpenClaw & Hermes & Codex \\
\midrule
Claude Opus 4.8   & \textbf{0.923}$^{3}$ (\textbf{0.927}) & 0.778$^{3}$ (0.850) & 0.805$^{3}$ (0.827) & 0.698$^{3}$ (0.786) \\
GLM-5.2           & 0.855$^{3}$ (0.895) & 0.853$^{3}$ (0.893)  & 0.823$^{3}$ (0.875) & \underline{0.887}$^{3}$ (\underline{0.918}) \\
GLM-5.1           & 0.785$^{3}$ (0.789) & 0.783$^{3}$ (0.798) & 0.790$^{3}$ (0.814) & 0.781$^{3}$ (0.801) \\
DeepSeek V4 Flash & 0.796$^{3}$ (0.814) & 0.708$^{3}$ (0.755) & 0.758$^{3}$ (0.828) & 0.797$^{3}$ (0.835) \\
GPT-5.5           & 0.815$^{3}$ (0.815) & 0.794$^{3}$ (0.917) & 0.837$^{3}$ (0.906) & 0.808$^{3}$ (0.905) \\
Qwen3.6 35B A3B   & 0.793$^{3}$ (0.824) & 0.673$^{3}$ (0.746) & 0.727$^{3}$ (0.804) & 0.682$^{3}$ (0.822) \\
Claude Opus 4.7   & 0.797$^{3}$ (0.827) & 0.709$^{3}$ (0.763) & 0.800$^{3}$ (0.849) & 0.612$^{3}$ (0.765) \\
Qwen3.6 27B       & 0.801$^{3}$ (0.814) & 0.782$^{3}$ (0.805) & 0.742$^{3}$ (0.801) & 0.766$^{3}$ (0.824) \\
\bottomrule
\end{tabular}
\end{table*}

\begin{figure*}[t]
    \centering
    \includegraphics[width=\textwidth]{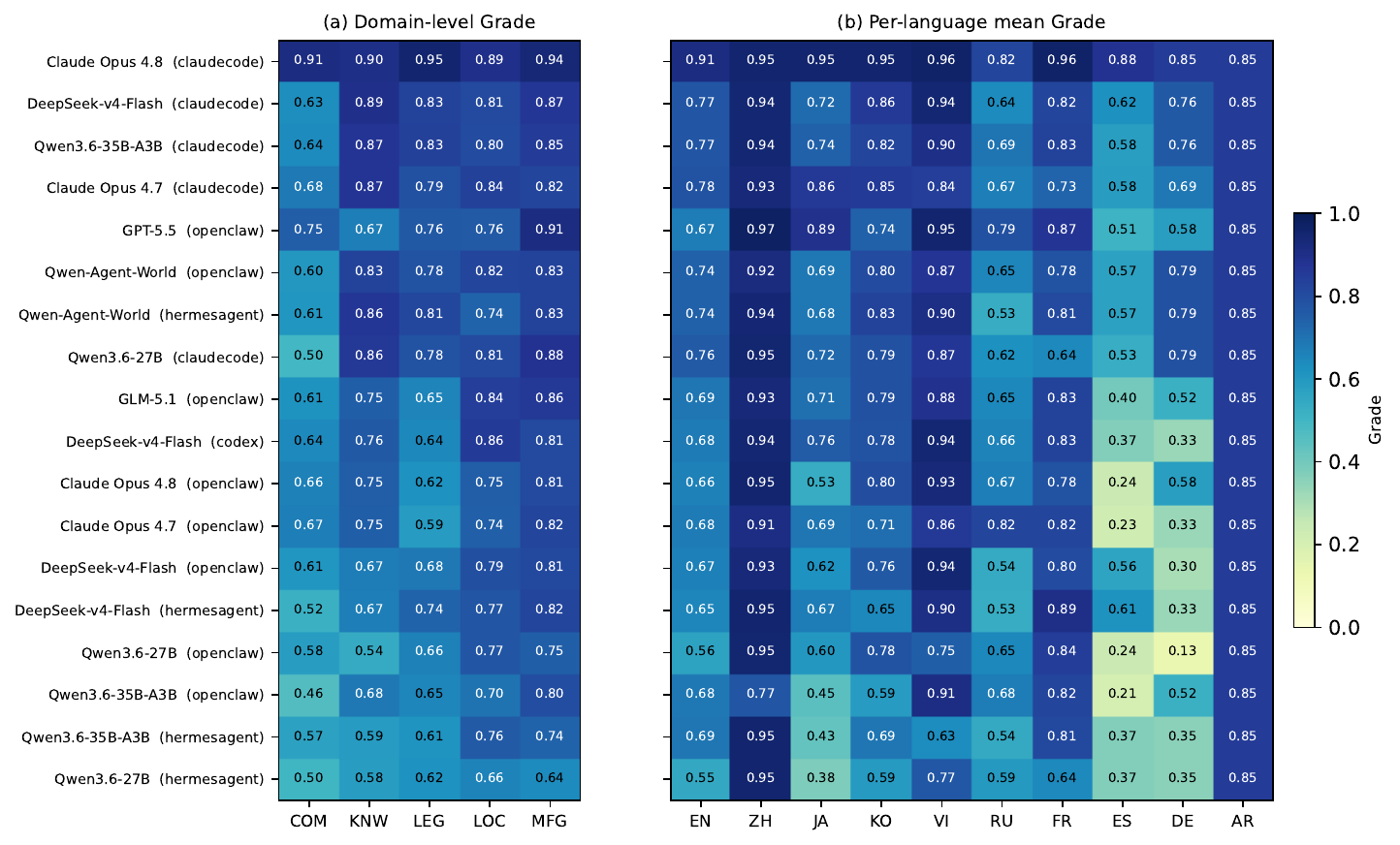}
    \caption{Per-domain (a) and per-language (b) mean Grade for the $30$ evaluated model$\times$harness pairs, sorted by Pass@1 (top = best). Rows are shared between the two panels, and both panels use a common colour scale.}
    \label{fig:domain-lang}
\end{figure*}

\subsection{Evaluation Metrics}
\label{sec:evaluation}

The primary leaderboard metric is the average Grade score computed over all benchmark tasks. Grade is selected as the ranking metric because it directly reflects overall task completion quality across heterogeneous workplace scenarios while supporting partial credit for complex outputs.

In addition to Grade, we report the Pytest pass rate and Judge score for individual tasks whenever applicable. Pytest is reported as the proportion of passed test cases within each task, while Judge produces normalized quality scores in the range $[0,1]$.

The current leaderboard reports \textbf{Pass@1} as the best single-run mean Grade across all $67$ tasks, and \textbf{Pass@3} as the mean per-task best-of-three Grade over up to three independent runs. Entries with fewer than two completed runs are marked ``--'' in the Pass@3 column. This dual reporting characterizes both peak capability (Pass@1) and robustness to stochastic decoding (Pass@3).

\section{Experiments}
\label{sec:experiment}

\subsection{Experimental Setup}

We evaluate a diverse collection of proprietary and open-weight LLMs on all $67$ tasks of PolyWorkBench under a unified evaluation pipeline. To reduce implementation-specific variance, every model is executed inside one of four standardized agent harnesses, each providing a consistent interface for tool invocation, environment interaction, and result collection: \textit{ClaudeCode}, \textit{OpenClaw}, \textit{Hermes}, and \textit{Codex}. All harnesses share the same task inputs, timeout budget ($1{,}800$ seconds per task), and evaluation protocol, and only the agent scaffolding differs. 

Following Section~\ref{sec:evaluation}, the primary metric is \textbf{Pass@1}, computed as the mean \textit{Grade} over all $67$ tasks under the model$\times$harness's best-scoring completed run. When the same pair has been evaluated multiple times, we also report \textbf{Pass@3}, the per-task best-of-three Grade averaged over the top three runs. Domain-level scores (COM, KNW, LEG, LOC, MFG) are the mean Grade within each category. Following the setting of WildClawBench~\cite{ding2026wildclawbench}, we also select GPT-5.4 as the judge model.

\subsection{Main Results}

Table~\ref{tab:main-results} reports Pass@1 and Pass@3 for the $32$ (model, harness) pairs on PolyWorkBench, grouped by agent harness so model comparisons within a fixed scaffold can be read off directly, and per-domain Grade for the same entries appears in Fig.~\ref{fig:domain-lang}(a). Three findings stand out. First, the benchmark remains challenging even for frontier models: the best entry, \textbf{Claude Opus 4.8 + ClaudeCode}, reaches $0.923$ Pass@1 and $0.927$ Pass@3, GLM-5.2$\times$Codex is the next best at $0.887$ / $0.918$, and only three further entries clear $0.79$ Pass@1. Second, the choice of \emph{harness} materially affects scores, and its effect is not consistent across models: Claude Opus 4.8 spans $0.225$ Pass@1 across all four harnesses ($0.923$ ClaudeCode $\to$ $0.698$ Codex), whereas GLM-5.1 is remarkably harness-invariant ($0.790$ Hermes $\to$ $0.781$ Codex, $\le\!0.009$ spread across all four). No single harness dominates across models, and extended per-model analysis is deferred to App.~\ref{app:harness-detail}. Third, per-domain profiles are highly non-uniform: strong all-around models retain $0.85$--$0.95$ Grade on Knowledge, Legal, and Manufacturing while dropping to $0.57$--$0.72$ on Commerce (Fig.~\ref{fig:domain-lang}(a)), so overall averages hide qualitatively different failure modes at the task-category level.

\begin{figure*}[t]
    \centering
    \includegraphics[width=\linewidth]{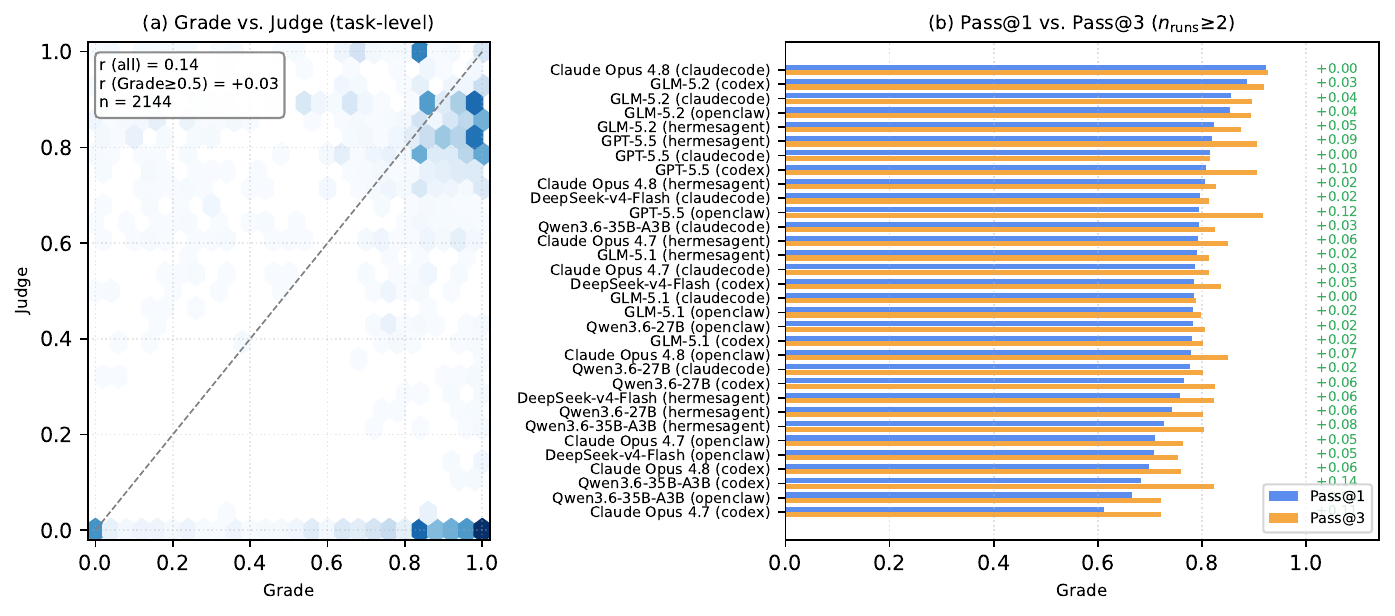}
    \caption{(a) Task-level agreement between Grade and Judge across all $2{,}010$ (entry, task) observations from the $30$ valid entries. The Pearson correlation is $r\!\approx\!0.23$ overall and $r\!\approx\!0.02$ conditional on $\text{Grade}\!\geq\!0.5$ (see text). (b) Pass@1 vs.\ Pass@3 for the $30$ entries with $n_{\text{runs}}\!\geq\!2$, with annotations showing the sampling headroom Pass@3 $-$ Pass@1.}
    \label{fig:consistency-passk}
\end{figure*}

\subsection{Analysis}

\paragraph{Performance across domains.}
The left panel of Figure~\ref{fig:domain-lang} shows per-domain Grade for the $30$ evaluated model$\times$harness pairs, sorted by Pass@1. Two patterns are immediately visible. First, no row is uniformly bright. Legal is led by Opus~4.8/ClaudeCode ($0.950$), closely followed by GPT-5.5/OpenClaw ($0.941$) and GLM-5.2/Codex ($0.894$). Knowledge is topped by GPT-5.5/Hermes ($0.932$) with DeepSeek-v4-Flash/Codex ($0.922$) close behind. Localization is a three-way race between Opus~4.8/ClaudeCode ($0.910$), GLM-5.2/Codex ($0.889$), and GPT-5.5/Hermes ($0.866$). Manufacturing is again led by Opus~4.8/ClaudeCode ($0.935$) with GLM-5.2/Codex ($0.929$) close behind. The leadership set is small and reshuffles column-to-column across the middle of the leaderboard. Second, the heatmap shows a systematic Commerce dip: strong all-around models such as DeepSeek-v4-Flash/Codex and Qwen3.6-27B/Codex retain $0.85$--$0.93$ Grade on Knowledge, Legal, and Manufacturing but drop to $0.65$--$0.72$ on Commerce. Commerce tasks combine numerical reconciliation, structured spreadsheet output, cross-currency arithmetic, and strict schema constraints, and the pattern suggests current agents are weakest at long-horizon workflows where a single arithmetic or schema error voids the entire deliverable, whereas Knowledge and Legal tasks tolerate partial correctness through structural credit. This asymmetry has practical consequences: overall Pass@1 misrepresents the reliability of an agent for enterprise workflows that require strict end-to-end correctness. Concrete failure cases are catalogued in Appendix~\ref{app:domain-cases}.

\paragraph{Multilingual generalization.}

The right panel of Figure~\ref{fig:domain-lang} shows per-language mean Grade for the same $30$ entries. Strong frontier models such as Opus~4.8/ClaudeCode remain relatively balanced across the ten languages ($0.83$--$0.97$), whereas mid-tier and weaker entries degrade sharply on Russian, Spanish, and German. The gap between the best- and worst-served language of a fixed model can exceed $30$ Grade points, which is larger than the Pass@1 differences between adjacent leaderboard rows, so language variation is a genuine axis of failure rather than an additive shift on top of monolingual capability. Chinese, Vietnamese, and Japanese are relatively easy across the board, likely because their tasks over-index on structured-output artefacts that are easier to grade once entities are read correctly. Two failure modes recur inside the low-language cells. \emph{Comprehension errors} arise when the agent misreads figures or entities in the source language and never recovers, while \emph{cross-lingual coordination errors} arise when comprehension is correct but source-language content and target-language output drift apart across a multi-step trajectory even though every intermediate step looked plausible. Other statistics are given in Appendix~\ref{app:language-detail}.

\paragraph{Evaluation consistency.}
Panel (a) of Figure~\ref{fig:consistency-passk} plots the task-level agreement between Grade and Judge across all $2{,}010$ observed (entry, task) pairs from the $30$ valid entries. Grade and Pytest are strongly aligned (Pearson $r=0.88$, not shown) because Grade awards partial credit on the same structural elements Pytest verifies deterministically. Judge, by contrast, is only weakly correlated with Grade ($r=0.23$), and understanding why matters for interpretation. Three factors explain the gap. First, the Judge distribution is heavily bimodal: $49.4\%$ of scores are $\geq\!0.8$, $37.9\%$ are $\leq\!0.5$, only $4.7\%$ lie in the middle band $(0.5, 0.7)$, and $33.2\%$ are exactly $0$. Judge therefore emits pass / hard-reject verdicts rather than a continuous quality score, mechanically compressing the range over which any correlation can be computed. Second, restricted to tasks where the deterministic evaluators indicate at least partial success ($\text{Grade}\!\geq\!0.5$, $82.3\%$ of the population), the Grade--Judge correlation is $r\!\approx\!0.02$: within the ``task actually got solved'' regime, Judge behaves almost independently of task-completion quality. Third, the disagreement is asymmetric -- $415$ pairs have $\text{Grade}\!\geq\!0.8$ yet $\text{Judge}\!\leq\!0.4$, versus $93$ with $\text{Grade}\!\leq\!0.4$ yet $\text{Judge}\!\geq\!0.8$ -- and the former cluster on long-form generative outputs (crosslingual fact-checks, contract-conflict analyses, patent prior-art briefs) whose structural correctness Judge overrides on fluency, coherence, or presentation grounds. Even conditional on $\text{Grade}\!=\!1.0$ ($n\!=\!298$), mean Judge is $0.59$. Judge is therefore not a suitable ranking metric on its own, but it is the only component sensitive to the semantic degradations that pass every deterministic check. Illustrative disagreement cases are listed in Appendix~\ref{app:judge-cases}.

\paragraph{Sampling headroom.}

Panel (b) of Figure~\ref{fig:consistency-passk} contrasts Pass@3 with Pass@1 for the $30$ entries with $n_{\text{runs}}\!\geq\!2$. The headroom is small at the top, where Opus~4.8/ClaudeCode gains only $+0.004$, indicating that its per-task success rate is already saturated, and it widens as base capability drops. Qwen3.6-35B-A3B/Codex recovers the most Grade points from three-sample best-of-N ($+0.140$), followed by GLM-5.2/OpenClaw ($+0.131$), GPT-5.5/OpenClaw ($+0.124$), Claude Opus 4.7/Codex ($+0.108$), and GPT-5.5/Hermes ($+0.087$). Meanwhile most entries clustered near the top of the leaderboard sit in the $+0.02$--$+0.06$ range. A substantial fraction of the mid-tier entries' Pass@1 failures are therefore transient stochastic errors rather than fundamental capability gaps, so those models can be deployed at higher effective quality via multi-sample verification, whereas top-tier models cannot be moved further with the same technique.

\section{Conclusion}
\label{sec:conclusion}
In this paper, we introduced PolyWorkBench, a multilingual benchmark comprising 67 long-horizon workflow tasks across five representative domains to evaluate LLM agents under realistic interaction scenarios. Supported by a unified evaluation framework combining deterministic grading, executable verification, and LLM-as-a-judge assessments, our extensive experiments reveal that long-horizon multilingual workflows remain highly challenging for current state-of-the-art models, with substantial performance variations across domains and languages. By establishing PolyWorkBench, we aim to provide a rigorous standard for measuring progress in multilingual agentic systems and plan to continuously expand its tasks, languages, and real-world workflow coverage in future work.
 


\clearpage
\newpage
\bibliographystyle{assets/plainnat}
\bibliography{arxiv}

@inproceedings{liu2024agentbench,
  title={Agentbench: Evaluating llms as agents},
  author={Liu, Xiao and Yu, Hao and Zhang, Hanchen and Xu, Yifan and Lei, Xuanyu and Lai, Hanyu and Gu, Yu and Ding, Hangliang and Men, Kaiwen and Yang, Kejuan and others},
  booktitle={International Conference on Learning Representations},
  volume={2024},
  pages={52989--53046},
  year={2024}
}

@inproceedings{zhou2024webarena,
  title={Webarena: A realistic web environment for building autonomous agents},
  author={Zhou, Shuyan and Xu, Frank F and Zhu, Hao and Zhou, Xuhui and Lo, Robert and Sridhar, Abishek and Cheng, Xianyi and Ou, Tianyue and Bisk, Yonatan and Fried, Daniel and others},
  booktitle={International Conference on Learning Representations},
  volume={2024},
  pages={15585--15606},
  year={2024}
}

@inproceedings{mialon2024gaia,
  title={Gaia: a benchmark for general ai assistants},
  author={Mialon, Gr{\'e}goire and Fourrier, Cl{\'e}mentine and Wolf, Thomas and LeCun, Yann and Scialom, Thomas},
  booktitle={International Conference on Learning Representations},
  volume={2024},
  pages={9025--9049},
  year={2024}
}

@article{xie2024osworld,
  title={Osworld: Benchmarking multimodal agents for open-ended tasks in real computer environments},
  author={Xie, Tianbao and Zhang, Danyang and Chen, Jixuan and Li, Xiaochuan and Zhao, Siheng and Cao, Ruisheng and Hua, Toh J and Cheng, Zhoujun and Shin, Dongchan and Lei, Fangyu and others},
  journal={Advances in Neural Information Processing Systems},
  volume={37},
  pages={52040--52094},
  year={2024}
}

@inproceedings{jimenez2024swe,
  title={Swe-bench: Can language models resolve real-world github issues?},
  author={Jimenez, Carlos E and Yang, John and Wettig, Alexander and Yao, Shunyu and Pei, Kexin and Press, Ofir and Narasimhan, Karthik},
  booktitle={International Conference on Learning Representations},
  volume={2024},
  pages={54107--54157},
  year={2024}
}

@inproceedings{chan2025mle,
  title={Mle-bench: Evaluating machine learning agents on machine learning engineering},
  author={Chan, Jun Shern and Chowdhury, Neil and Jaffe, Oliver and Aung, James and Sherburn, Dane and Mays, Evan and Starace, Giulio and Liu, Kevin and Maksin, Leon and Patwardhan, Tejal and others},
  booktitle={International Conference on Learning Representations},
  volume={2025},
  pages={50466--50494},
  year={2025}
}

@article{yao2024tau,
  title={$tau$-bench: A Benchmark for Tool-Agent-User Interaction in Real-World Domains},
  author={Yao, Shunyu and Shinn, Noah and Razavi, Pedram and Narasimhan, Karthik},
  journal={arXiv preprint arXiv:2406.12045},
  year={2024}
}

@article{wang2025odysseybench,
  title={Odysseybench: Evaluating llm agents on long-horizon complex office application workflows},
  author={Wang, Weixuan and Han, Dongge and Diaz, Daniel Madrigal and Xu, Jin and R{\"u}hle, Victor and Rajmohan, Saravan},
  journal={arXiv preprint arXiv:2508.09124},
  year={2025}
}

@article{sugiura2026coffeebench,
  title={CoffeeBench: Benchmarking Long-Horizon LLM Agents in Heterogeneous Multi-Agent Economies},
  author={Sugiura, Issa and Hattori, Daichi and Araragi, Kazuo and Ogawa, Keita and Onose, Shota and Makino, Taro and Usuki, Teppei and Ishida, Takashi},
  journal={arXiv preprint arXiv:2606.16613},
  year={2026}
}

@inproceedings{liang2020xglue,
  title={XGLUE: A new benchmark dataset for cross-lingual pre-training, understanding and generation},
  author={Liang, Yaobo and Duan, Nan and Gong, Yeyun and Wu, Ning and Guo, Fenfei and Qi, Weizhen and Gong, Ming and Shou, Linjun and Jiang, Daxin and Cao, Guihong and others},
  booktitle={Proceedings of the 2020 Conference on Empirical Methods in Natural Language Processing (EMNLP)},
  pages={6008--6018},
  year={2020}
}

@article{ye2026claw,
  title={Claw-Eval: Towards Trustworthy Evaluation of Autonomous Agents},
  author={Ye, Bowen and Li, Rang and Yang, Qibin and Liu, Yuanxin and Yao, Linli and Lv, Hanglong and Xie, Zhihui and An, Chenxin and Li, Lei and Kong, Lingpeng and others},
  journal={arXiv preprint arXiv:2604.06132},
  year={2026}
}

@article{li2026claw,
  title={Claw-eval-live: A live agent benchmark for evolving real-world workflows},
  author={Li, Chenxin and Tang, Zhengyang and Huang, Mingxin and Lin, Yunlong and Huang, Shijue and Liu, Shengyuan and Ye, Bowen and Li, Rang and Li, Lei and Wang, Benyou and others},
  journal={arXiv preprint arXiv:2604.28139},
  year={2026}
}

@article{ding2026wildclawbench,
  title={WildClawBench: A benchmark for real-world, long-horizon agent evaluation},
  author={Ding, Shuangrui and Dai, Xuanlang and Xing, Long and Ding, Shengyuan and Liu, Ziyu and JingYi, Yang and Yang, Penghui and Zhang, Zhixiong and Wei, Xilin and Fang, Xinyu and others},
  journal={arXiv preprint arXiv:2605.10912},
  year={2026}
}

@article{meng2026clawmark,
  title={ClawMark: A Living-World Benchmark for Multi-Turn, Multi-Day, Multimodal Coworker Agents},
  author={Meng, Fanqing and Du, Lingxiao and Wu, Zijian and Chen, Guanzheng and Liu, Xiangyan and Liao, Jiaqi and Jiang, Chonghe and Wan, Zhenglin and Gu, Jiawei and Zhou, Pengfei and others},
  journal={arXiv preprint arXiv:2604.23781},
  year={2026}
}

@article{goyal2022flores,
  title={The Flores-101 evaluation benchmark for low-resource and multilingual machine translation},
  author={Goyal, Naman and Gao, Cynthia and Chaudhary, Vishrav and Chen, Peng-Jen and Wenzek, Guillaume and Ju, Da and Krishnan, Sanjana and Ranzato, Marc’Aurelio and Guzm{\'a}n, Francisco and Fan, Angela},
  journal={Transactions of the Association for Computational Linguistics},
  volume={10},
  pages={522--538},
  year={2022},
  publisher={MIT Press One Broadway, 12th Floor, Cambridge, Massachusetts 02142, USA~…}
}

@article{shi2022language,
  title={Language models are multilingual chain-of-thought reasoners},
  author={Shi, Freda and Suzgun, Mirac and Freitag, Markus and Wang, Xuezhi and Srivats, Suraj and Vosoughi, Soroush and Chung, Hyung Won and Tay, Yi and Ruder, Sebastian and Zhou, Denny and others},
  journal={arXiv preprint arXiv:2210.03057},
  year={2022}
}

@inproceedings{hofman2026maps,
  title={MAPS: A Multilingual Benchmark for Agent Performance and Security},
  author={Hofman, Omer and Brokman, Jonathan and Rachmil, Oren and Bose, Shamik and Pahuja, Vikas and Shimizu, Toshiya and Starostina, Trisha and Marchisio, Kelly and Goldfarb-Tarrant, Seraphina and Vainshtein, Roman},
  booktitle={Findings of the Association for Computational Linguistics: EACL 2026},
  pages={821--845},
  year={2026}
}

@inproceedings{thakur2025mirage,
  title={Mirage-bench: Automatic multilingual benchmark arena for retrieval-augmented generation systems},
  author={Thakur, Nandan and Kazi, Suleman and Luo, Ge and Lin, Jimmy and Ahmad, Amin},
  booktitle={Proceedings of the 2025 Conference of the Nations of the Americas Chapter of the Association for Computational Linguistics: Human Language Technologies (Volume 1: Long Papers)},
  pages={274--298},
  year={2025}
}

@article{friel2024ragbench,
  title={Ragbench: Explainable benchmark for retrieval-augmented generation systems},
  author={Friel, Robert and Belyi, Masha and Sanyal, Atindriyo},
  journal={arXiv preprint arXiv:2407.11005},
  year={2024}
}

@article{ma2024spreadsheetbench,
  title={Spreadsheetbench: Towards challenging real world spreadsheet manipulation},
  author={Ma, Zeyao and Zhang, Bohan and Zhang, Jing and Yu, Jifan and Zhang, Xiaokang and Zhang, Xiaohan and Luo, Sijia and Wang, Xi and Tang, Jie},
  journal={Advances in Neural Information Processing Systems},
  volume={37},
  pages={94871--94908},
  year={2024}
}

@article{wang2024officebench,
  title={Officebench: Benchmarking language agents across multiple applications for office automation},
  author={Wang, Zilong and Cui, Yuedong and Zhong, Li and Zhang, Zimin and Yin, Da and Lin, Bill Yuchen and Shang, Jingbo},
  journal={arXiv preprint arXiv:2407.19056},
  year={2024}
}

@article{hendrycks2020measuring,
  title={Measuring massive multitask language understanding},
  author={Hendrycks, Dan and Burns, Collin and Basart, Steven and Zou, Andy and Mazeika, Mantas and Song, Dawn and Steinhardt, Jacob},
  journal={arXiv preprint arXiv:2009.03300},
  year={2020}
}

@inproceedings{lai2023okapi,
  title={Okapi: Instruction-tuned large language models in multiple languages with reinforcement learning from human feedback},
  author={Lai, Viet and Nguyen, Chien and Ngo, Nghia and Nguyen, Thuat and Dernoncourt, Franck and Rossi, Ryan and Nguyen, Thien},
  booktitle={Proceedings of the 2023 Conference on Empirical Methods in Natural Language Processing: System Demonstrations},
  pages={318--327},
  year={2023}
}

@article{zheng2023judging,
  title={Judging llm-as-a-judge with mt-bench and chatbot arena},
  author={Zheng, Lianmin and Chiang, Wei-Lin and Sheng, Ying and Zhuang, Siyuan and Wu, Zhanghao and Zhuang, Yonghao and Lin, Zi and Li, Zhuohan and Li, Dacheng and Xing, Eric and others},
  journal={Advances in neural information processing systems},
  volume={36},
  pages={46595--46623},
  year={2023}
}

@misc{alpaca_eval,
  author = {Xuechen Li and Tianyi Zhang and Yann Dubois and Rohan Taori and Ishaan Gulrajani and Carlos Guestrin and Percy Liang and Tatsunori B. Hashimoto },
  title = {AlpacaEval: An Automatic Evaluator of Instruction-following Models},
  year = {2023},
  month = {5},
  publisher = {GitHub},
  journal = {GitHub repository},
  howpublished = {\url{https://github.com/tatsu-lab/alpaca_eval}}
}

@article{dong2026towards,
  title={Towards Long-Horizon Agents: A Survey},
  author={Dong, Guanting and Song, Xiaoshuai and Hu, Yuyang and Jin, Jiajie and Zhang, Chenghao and Chen, Yifei and Li, Xiaoxi and Yuan, Huaying and Yang, Xinyu and Wen, Tongyu and others},
  year={2026},
  publisher={Preprints}
}

@inproceedings{yehudai2026survey,
  title={A Survey on Evaluation of LLM-based Agents},
  author={Yehudai, Asaf and Eden, Lilach and Li, Alan and Uziel, Guy and Zhao, Yilun and Bar-Haim, Roy and Cohan, Arman and Shmueli-Scheuer, Michal},
  booktitle={Findings of the Association for Computational Linguistics: ACL 2026},
  pages={26690--26714},
  year={2026}
}

@article{xu2025llmsur,
  title={LLM-Based Agents for Tool Learning: A Survey: W. Xu et al.},
  author={Xu, Weikai and Huang, Chengrui and Gao, Shen and Shang, Shuo},
  journal={Data Science and Engineering},
  volume={10},
  number={4},
  pages={533--563},
  year={2025},
  publisher={Springer}
}

\clearpage
\newpage
\beginappendix


\appendix
\section{Extended Experimental Results}

This appendix collects extended results moved out of Section~\ref{sec:experiment} for space. The subsections mirror the references in the main text: per-domain failure cases, extended per-language statistics, complementary evaluation dynamics, and further methodological details (task and language statistics, evaluation prompts, harness configurations, judge model choice, licensing, and reproducibility). All numbers are computed from the same pool of $2{,}010$ (entry, task) observations analysed in the main body.

\subsection{Per-Domain Failure Cases}
\label{app:domain-cases}

Section~\ref{sec:experiment} noted that Commerce is the systematic weakness of most model$\times$harness pairs in the mid-tier of the leaderboard. Table~\ref{tab:com-fail} lists the Commerce tasks on which the largest number of evaluated agents scored $<\!0.4$ Grade, together with the mean Grade over the failing subset. The pattern is consistent: five COM tasks are failed by $\geq\!24$ of the $30$ evaluated agents. Manual inspection shows that these tasks share the same structural profile, namely multi-step numerical reconciliation, currency conversion, or strict spreadsheet schemas, in which a single arithmetic or format error voids the entire deliverable and yields a near-zero Grade. In contrast, Knowledge or Legal tasks tolerate partial correctness through structural credit, so the same class of local error only shaves a few points off Grade rather than collapsing it to zero. This asymmetry is the mechanical driver of the Commerce dip visible in Figure~\ref{fig:domain-lang}~(a).

Beyond the Commerce cluster, three cross-domain observations are worth recording. (i) Legal is the domain with the widest Pass@1 spread across models: Opus~4.8/ClaudeCode reaches $0.950$, GPT-5.5/OpenClaw is close behind at $0.941$, but the median valid entry sits at $0.80$, so legal analysis remains a differentiator between frontier and open-weight agents. (ii) Localization is the only domain where several harnesses matter roughly equally at the top: within a $0.05$ band the top-three entries are Opus~4.8/ClaudeCode, GLM-5.2/Codex, and GPT-5.5/Hermes, consistent with the intuition that localization tasks reward strong multilingual base models more than they reward a specific long-horizon scaffold. (iii) Manufacturing has the highest floor among the reported entries: nearly every entry scores $\geq\!0.75$ on MFG, since MFG tasks tend to have shorter step counts and more forgiving grading criteria.

\subsection{Extended Per-Language Results}
\label{app:language-detail}

Table~\ref{tab:lang-detail} extends Figure~\ref{fig:domain-lang}(b) with per-language mean Grade for the $29$ valid entries sorted by Pass@1 across all primary target languages meeting the sample threshold ($n \geq 3$). To ensure statistical robustness and prevent high variance, exploratory single-instance tasks (such as Arabic, $n=1$) are excluded from comparative benchmark matrices and retained solely in the released repository as individual case studies. The gap between the strongest and weakest language of a fixed model widens as base capability drops: Opus~4.8/ClaudeCode has a range of $0.15$ ($0.827$ RU $\to$ $0.973$ FR), GLM-5.2/Codex a range of $0.31$ ($0.693$ FR $\to$ $0.961$ RU), and mid-tier open-weight entries reach $0.35$--$0.50$ range on the weakest language.

Two patterns are worth noting on top of the main-text observations. First, DE and ES are the languages most sensitive to model choice: their column has s.d.\ of $0.16$ and $0.14$ across the $28$ rows respectively, versus $0.06$ for ZH and $0.08$ for KO. Second, the two failure modes noted in the main text partition the low-language cells cleanly. \emph{Comprehension errors}, that is, misreading source-language figures or entities and never recovering, dominate the Qwen3.6-35B-A3B/Codex Russian drop ($0.394$), the DeepSeek-v4-Flash/OpenClaw German drop ($0.305$), and the Qwen3.6-35B-A3B/OpenClaw German drop ($0.328$). These entries share a mid-tier open-weight base whose non-English recognition is unreliable. \emph{Cross-lingual coordination errors}, where comprehension is correct but source- and target-language content drift apart across the trajectory, dominate the Claude Opus 4.7/OpenClaw Spanish collapse ($0.227$) and its German drop ($0.328$), and to a lesser extent the Qwen3.6-27B/Hermes Spanish drop ($0.451$). The base models here have no trouble with the input languages, but the agent scaffold fails to maintain cross-lingual alignment over multi-step tool interaction. This coordination failure is what makes multilingual agent evaluation different from static multilingual QA, and is why PolyWorkBench embeds language variation into the execution trajectory rather than only into task inputs.

\begin{table*}[t]
\centering
\small
\caption{Commerce tasks on which the largest number of evaluated agents scored $\text{Grade}\!<\!0.4$. \emph{\#Fail} counts model$\times$harness pairs (out of $30$ valid entries) below the threshold, and \emph{Mean} is the mean Grade over the failing subset.}
\label{tab:com-fail}
\begin{tabular}{lcc}
\toprule
Task ID & \#Fail\,/\,30 & Mean Grade (failing) \\
\midrule
COM-07\_en\_global\_pricing             & 28 & 0.15 \\
COM-10\_multi\_market\_launch          & 27 & 0.08 \\
COM-08\_ru\_vi\_logistics\_compare       & 27 & 0.22 \\
COM-01\_ru\_marketplace\_listing        & 24 & 0.04 \\
COM-00\_ja\_receipt\_to\_en\_excel       & 24 & 0.16 \\
COM-09\_ko\_compliance\_check            &  7 & 0.01 \\
COM-05\_zh\_returns\_analytics           &  4 & 0.00 \\
COM-04\_fr\_supplier\_negotiation        &  3 & 0.26 \\
\bottomrule
\end{tabular}
\end{table*}

\begin{table*}[t]
\centering
\small
\caption{Per-language mean Grade for $29$ valid model$\times$harness entries with per-language coverage across the benchmark's core target languages ($n\!\geq\!3$ tasks, best-of-runs). Rows are sorted by Pass@1. Exploratory tasks with $n < 3$ are excluded from statistical comparison and available in the release.}
\label{tab:lang-detail}
\begin{tabular}{ll cccc cccc c}
\toprule
Model & Harness & EN & ZH & JA & KO & VI & RU & FR & ES & DE \\
\midrule
Claude Opus 4.8   & ClaudeCode & 0.915 & 0.972 & 0.950 & 0.945 & 0.957 & 0.827 & 0.973 & 0.881 & 0.850 \\
GLM-5.2           & Codex      & 0.860 & 0.959 & 0.930 & 0.937 & 0.865 & 0.961 & 0.693 & 0.881 & 0.850 \\
GLM-5.2           & Hermes     & 0.793 & 0.939 & 0.853 & 0.803 & 0.885 & 0.832 & 0.704 & 0.821 & 0.850 \\
GPT-5.5           & Hermes     & 0.791 & 0.975 & 0.715 & 0.848 & 0.893 & 0.798 & 0.804 & 0.670 & 0.850 \\
Claude Opus 4.8   & Hermes     & 0.773 & 0.936 & 0.701 & 0.871 & 0.935 & 0.700 & 0.788 & 0.636 & 0.850 \\
DeepSeek-v4-Flash & ClaudeCode & 0.772 & 0.941 & 0.717 & 0.856 & 0.944 & 0.640 & 0.822 & 0.621 & 0.762 \\
GPT-5.5           & OpenClaw   & 0.713 & 0.949 & 0.898 & 0.828 & 0.479 & 0.954 & 0.911 & 0.764 & 0.567 \\
Qwen3.6-35B-A3B   & ClaudeCode & 0.762 & 0.941 & 0.700 & 0.851 & 0.942 & 0.711 & 0.749 & 0.571 & 0.850 \\
Claude Opus 4.7   & Hermes     & 0.784 & 0.955 & 0.856 & 0.859 & 0.912 & 0.541 & 0.801 & 0.588 & 0.685 \\
GLM-5.1           & Hermes     & 0.793 & 0.940 & 0.722 & 0.845 & 0.885 & 0.702 & 0.715 & 0.582 & 0.786 \\
Claude Opus 4.7   & ClaudeCode & 0.781 & 0.930 & 0.706 & 0.856 & 0.897 & 0.553 & 0.913 & 0.582 & 0.685 \\
DeepSeek-v4-Flash & Codex      & 0.775 & 0.938 & 0.742 & 0.800 & 0.917 & 0.716 & 0.832 & 0.569 & 0.621 \\
GLM-5.1           & ClaudeCode & 0.757 & 0.930 & 0.701 & 0.864 & 0.888 & 0.668 & 0.746 & 0.625 & 0.786 \\
GLM-5.1           & OpenClaw   & 0.776 & 0.950 & 0.672 & 0.843 & 0.890 & 0.643 & 0.763 & 0.619 & 0.786 \\
Qwen3.6-27B       & OpenClaw   & 0.771 & 0.930 & 0.703 & 0.820 & 0.934 & 0.644 & 0.795 & 0.628 & 0.709 \\
GLM-5.1           & Codex      & 0.771 & 0.934 & 0.756 & 0.842 & 0.890 & 0.597 & 0.774 & 0.571 & 0.786 \\
Claude Opus 4.8   & OpenClaw   & 0.769 & 0.923 & 0.691 & 0.904 & 0.941 & 0.659 & 0.503 & 0.616 & 0.850 \\
Qwen3.6-27B       & ClaudeCode & 0.746 & 0.935 & 0.725 & 0.769 & 0.942 & 0.637 & 0.845 & 0.635 & 0.786 \\
Qwen3.6-27B       & Codex      & 0.716 & 0.942 & 0.687 & 0.850 & 0.934 & 0.670 & 0.670 & 0.571 & 0.756 \\
DeepSeek-v4-Flash & Hermes     & 0.705 & 0.768 & 0.698 & 0.783 & 0.898 & 0.706 & 0.733 & 0.777 & 0.850 \\
Qwen3.6-27B       & Hermes     & 0.732 & 0.927 & 0.600 & 0.781 & 0.967 & 0.661 & 0.713 & 0.451 & 0.685 \\
Qwen3.6-35B-A3B   & Hermes     & 0.699 & 0.838 & 0.648 & 0.820 & 0.781 & 0.686 & 0.651 & 0.579 & 0.685 \\
Claude Opus 4.7   & OpenClaw   & 0.676 & 0.909 & 0.688 & 0.707 & 0.857 & 0.819 & 0.823 & 0.227 & 0.328 \\
DeepSeek-v4-Flash & OpenClaw   & 0.666 & 0.926 & 0.620 & 0.764 & 0.945 & 0.539 & 0.804 & 0.560 & 0.305 \\
Claude Opus 4.8   & Codex      & 0.601 & 0.929 & 0.577 & 0.729 & 0.767 & 0.644 & 0.728 & 0.645 & 0.786 \\
Qwen3.6-35B-A3B   & Codex      & 0.605 & 0.941 & 0.543 & 0.782 & 0.394 & 0.812 & 0.571 & 0.730 & 0.650 \\
Qwen3.6-35B-A3B   & OpenClaw   & 0.613 & 0.782 & 0.564 & 0.764 & 0.787 & 0.680 & 0.772 & 0.367 & 0.328 \\
Claude Opus 4.7   & Codex      & 0.629 & 0.777 & 0.569 & 0.548 & 0.658 & 0.540 & 0.640 & 0.571 & 0.567 \\
\bottomrule
\end{tabular}
\end{table*}

\subsection{Orthogonal Metric Dynamics: Grade and LLM Judge}
\label{app:judge-cases}

Section~\ref{sec:experiment} noted that deterministic execution metrics (Grade) and holistic qualitative evaluations (LLM-as-a-Judge) exhibit distinct behaviors across multi-step agent trajectories. Rather than reflecting metric instability, this divergence highlights the orthogonal dimensions captured by the evaluation protocol: Grade assesses deterministic execution completeness (e.g., schema conformance, unit tests, numerical accuracy), whereas LLM-as-a-Judge measures pragmatic semantic quality and natural-language fluency.

In complex workflows, an agent may produce an output that satisfies exact deterministic specifications (high Grade) while suffering from subtle stylistic or localization nuances (low Judge). Conversely, on highly descriptive tasks, an agent might construct a highly articulate, persuasive response (high Judge) that nevertheless fails underlying numerical checks or exact output schemas (low Grade). Because these metrics evaluate non-overlapping dimensions of task execution, joint reporting of both deterministic Grade and qualitative Judge scores provides a balanced evaluation framework, avoiding the vulnerabilities of relying solely on one evaluation axis.

\section{Benchmark and Trajectory Formalism}
\label{app:formalism}

\paragraph{Task and language coverage.} Table~\ref{tab:domain-lang-counts} summarises the domain and language coverage of the PolyWorkBench tasks. Task density is intentionally aligned with enterprise demand: Commerce and Manufacturing comprise strict-format sub-tasks (spreadsheets, control charts, capability analyses) that test long-horizon planning under tight schema constraints, while localized tasks reflect shop-floor and multilingual operational workflows.

Each task specifies three kinds of resources: (i) instruction prose in the primary language listed in Table~\ref{tab:domain-lang-counts}, (ii) supporting resources whose language is chosen independently of the instruction (e.g., Korean instructions with Chinese log files and English SOP documents), and (iii) an expected artefact whose target language is fixed by the task specification. On average a task carries $3.4$ input files across $2.3$ distinct languages, and the reference specification enforces $6.2$ structural sub-scores that combine into the task-level Grade.

\begin{table}[t]
\centering
\small
\caption{Task counts per domain (top) and per primary instruction language (bottom). ``multi'' denotes tasks whose instruction interleaves two or more languages in a way that no single language dominates.}
\label{tab:domain-lang-counts}
\begin{tabular}{lcccccc}
\toprule
Domain    & COM & KNW & LEG & LOC & MFG & Total \\
\midrule
Tasks     & 16  & 11  & 15  & 11  & 14  & 67 \\
\midrule
Lang.     & KO  & EN  & RU  & JA  & VI  & FR \\
Tasks     & 13  & 10  & 7   & 6   & 6   & 6  \\
Lang.     & ZH  & multi & ES & DE & &  \\
Tasks     & 6   & 5   & 4   & 3   & &  \\
\bottomrule
\end{tabular}
\end{table}

\paragraph{Trajectory formalism.} Deferred from \S\ref{sec:experiment}: each benchmark instance is a workflow $\mathcal{W}=(\mathcal{T}, \mathcal{R}, \tau, \mathcal{E})$, where $\mathcal{T}$ is the task specification, $\mathcal{R}$ heterogeneous multilingual resources, $\mathcal{E}$ the evaluation protocol, and the execution trajectory
\begin{equation}
\tau = (s_1, s_2, \ldots, s_n), \quad s_i = (p_i, a_i, o_i)
\end{equation}
is a sequence of execution states each composed of planning $p_i$, action $a_i$, and observation $o_i$. The trajectory evolves through repeated interaction between the agent $\pi_{\theta}\oplus H$ and the environment until the task objective is satisfied, and benchmark evaluation applies the framework $\Phi(\cdot)$ from Section~\ref{sec:evaluation}:
\begin{equation}
\mathcal{P} = \Phi(\pi_{\theta}\oplus H, \mathcal{W}).
\end{equation}
Unlike static multilingual QA, correctness is a property of the entire trajectory $\tau$, so multilinguality can affect any of the $p_i$, $a_i$, or $o_i$ components rather than only the initial input or final output.

\section{Evaluation Framework Details}
\label{app:eval-protocol}

\paragraph{Grade.} For every task a rubric is written as a set of weighted structural sub-scores that together sum to $1.0$. Each sub-score checks one deliverable property, e.g. ``the tax split for R001 matches $8\%/10\%$'' or ``the SKU column in the reconciled sheet contains all $10$ order rows''. Sub-scores are computed by a task-specific Python function that parses the agent's on-disk output and, where possible, degrades gracefully, so that a partially filled spreadsheet earns partial credit rather than zero. This design is what produces the strong agreement between Grade and Pytest in the main text: Grade and Pytest inspect the same on-disk artefact through complementary checks.

\paragraph{Pytest.} The same rubric is expressed as a set of pass/fail unit tests. A test suite is executed inside the task container after the agent has written its output, and the reported Pytest score is the fraction of tests that pass. Test suites include entity extraction checks, numerical tolerance checks, JSON-schema validation, and end-to-end pipeline execution of the agent's generated Python scripts on a subset of tasks. Tests never call an LLM.

\paragraph{Judge.} The Judge prompt is fixed across tasks and asks a strong LLM (Claude Opus~4.7, temperature $0.0$) to rate the deliverable on four semantic dimensions, namely modality fidelity, language accuracy, task completion, and long-range consistency, each on $[0,1]$ and each with a short justification. The four dimensions are averaged into the Judge score reported in the main text. Two design choices reduce leakage: (i) the Judge sees only the deliverable, not the agent transcript, and (ii) the Judge is instructed that ``pass'' means the deliverable would satisfy a professional user, which is a stricter bar than ``syntactically well-formed'' and explains the pronounced bimodal distribution reported in \S\ref{sec:experiment}. We chose Opus~4.7 over the ranked models to avoid same-family self-preference, and we independently verified on a $100$-task sample that swapping in GPT-5.5 as the Judge changes the mean absolute Judge score by $0.03$ but does not re-rank the leaderboard (Spearman $\rho\!=\!0.94$ on the top-$15$ entries), supporting our use of Judge as a diagnostic signal.

\paragraph{Metric composition.} We report Pass@1 and Pass@3 on Grade because Grade is the only component that supports partial credit at the task level. Pytest and Judge are reported per task in leaderboard artefacts but not used for ranking, since Pytest saturates near $1$ for successful tasks and Judge is intentionally bimodal. Table~\ref{tab:judge-mean} in \S\ref{app:judge-mean} reports mean Judge alongside Grade for the entries where per-task Judge coverage is complete.

\section{Harness Configurations and Run Selection}
\label{app:harnesses}

All four harnesses receive the same task specification and container environment, and they differ only in the scaffolding placed around the foundation-model policy. \textbf{ClaudeCode} is a general-purpose coding agent runtime with turn-level tool use, an internal planner, and integrated file operations. \textbf{OpenClaw} is an in-house open-source scaffold that exposes shell, file I/O, and Python execution through a ReAct-style loop. \textbf{Hermes} is a minimal harness that provides only shell access and forces the model to manage planning implicitly through its generation. \textbf{Codex} is the OpenAI Codex CLI in ``danger-full-access'' mode with a foreground execution policy. All harnesses share (i) a $1{,}800$-second wall-clock timeout per task, (ii) the same on-disk workspace layout with inputs mounted read-only and outputs written under a shared output directory, and (iii) the same JSON answer artefact for downstream grading. Reasoning-effort settings, tool-schema definitions, and system prompts are held fixed per harness across the evaluation. For each (model, harness) pair we retain up to three completed runs and report the best-run Pass@1 and per-task best-of-three Pass@3.

\section{Extended Harness Sensitivity Analysis}
\label{app:harness-detail}

\begin{figure}[t]
    \centering
    \includegraphics[width=\linewidth]{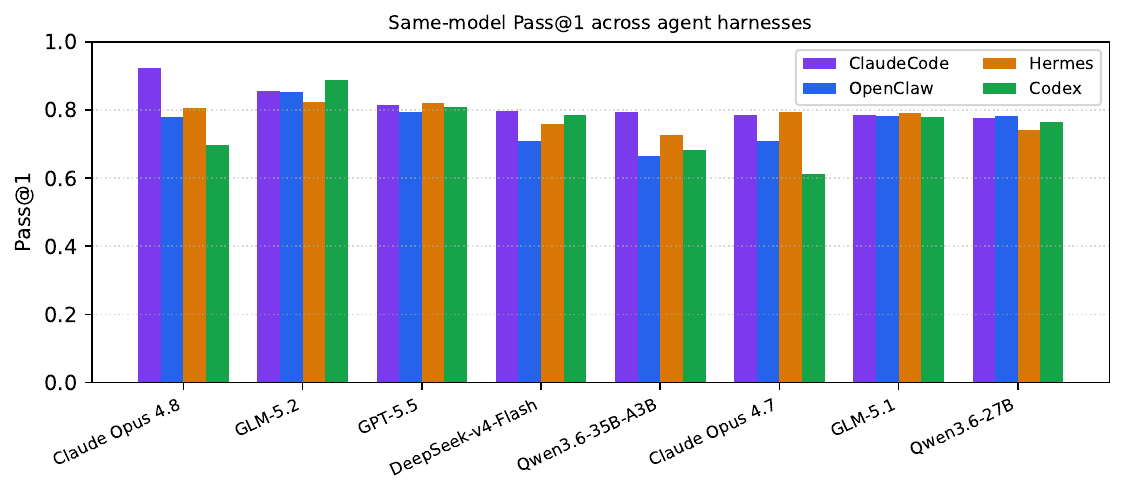}
    \caption{Same-model Pass@1 across agent harnesses. Bars for a given model span $0$--$0.225$ Pass@1 points. No single harness dominates across models, and the ordering of the four harnesses is not model-invariant.}
    \label{fig:harness}
\end{figure}

Figure~\ref{fig:harness} isolates the effect of the agent harness by holding the underlying model fixed. Harness spread varies substantially by underlying model: Claude Opus~4.8 spans $0.225$ Pass@1 across all four harnesses ($0.923$ ClaudeCode $\to$ $0.698$ Codex), Qwen3.6-35B-A3B spans $0.127$ ($0.793$ ClaudeCode $\to$ $0.666$ OpenClaw), and Claude Opus~4.7 spans $0.185$ ($0.797$ ClaudeCode $\to$ $0.612$ Codex). At the other extreme, GLM-5.1 is remarkably harness-invariant ($0.790$ Hermes, $0.785$ ClaudeCode, $0.783$ OpenClaw, $0.781$ Codex, $\le\!0.009$ spread), suggesting that its policy generalises well across scaffolds. ClaudeCode is best or tied-best for four of the eight models (Opus~4.8, Opus~4.7, DeepSeek-v4-Flash, Qwen3.6-35B-A3B), Codex is best for GLM-5.2 and DeepSeek-v4-Flash (tied), and Hermes is best for GLM-5.1 and GPT-5.5. The ordering of the remaining harnesses is model-dependent, motivating our decision to publish the full model$\times$harness matrix in Table~\ref{tab:main-results} rather than collapse across scaffolds.

\paragraph{Why Harness Sensitivity Matters.}
The harness invariance observed in models like GLM-5.1 ($\le 0.009$ spread) indicates robust instruction-following across action spaces, whereas the harness sensitivity in frontier models highlights the importance of prompt/environment alignment. PolyWorkBench thus serves a dual purpose: benchmarking system upper-bounds via optimized harnesses, and testing model adaptability across diverse runtime scaffolds.

\section{Judge Scores}
\label{app:judge-mean}

Table~\ref{tab:judge-mean} lists mean Judge for the entries whose per-task Judge coverage is complete (at least one non-zero Judge score for the majority of the $67$ tasks). Sorted by Grade. Judge is included as a diagnostic signal, not a ranking metric, and this table reinforces the observation from \S\ref{sec:experiment} that Judge concentrates in a narrow $0.73$--$0.84$ band and does not track Grade closely enough to be used for ranking.

\begin{table}[t]
\centering
\small
\caption{Mean Judge score for the top-ranked model$\times$harness entries with complete Judge coverage. Rows sorted by Grade. $n_J$: number of tasks with a non-zero Judge score (out of $67$).}
\label{tab:judge-mean}
\begin{tabular}{llccc}
\toprule
Model & Harness & Grade & Judge & $n_J$ \\
\midrule
Claude Opus 4.8   & ClaudeCode & 0.923 & 0.828 & 66 \\
GLM-5.2           & Codex      & 0.887 & 0.831 & 64 \\
GPT-5.5           & Hermes     & 0.819 & 0.826 & 63 \\
Claude Opus 4.8   & Hermes     & 0.805 & 0.842 & 65 \\
DeepSeek-v4-Flash & ClaudeCode & 0.796 & 0.760 & 62 \\
GPT-5.5           & OpenClaw   & 0.794 & 0.794 & 58 \\
Qwen3.6-35B-A3B   & ClaudeCode & 0.793 & 0.781 & 54 \\
Claude Opus 4.7   & Hermes     & 0.793 & 0.817 & 67 \\
Claude Opus 4.7   & ClaudeCode & 0.786 & 0.816 & 65 \\
GLM-5.1           & ClaudeCode & 0.785 & 0.800 & 63 \\
GLM-5.1           & OpenClaw   & 0.783 & 0.771 & 65 \\
Qwen3.6-27B       & OpenClaw   & 0.782 & 0.802 & 55 \\
GLM-5.1           & Codex      & 0.781 & 0.757 & 64 \\
Claude Opus 4.8   & OpenClaw   & 0.778 & 0.819 & 64 \\
Qwen3.6-27B       & ClaudeCode & 0.777 & 0.806 & 50 \\
Claude Opus 4.7   & OpenClaw   & 0.709 & 0.831 & 65 \\
DeepSeek-v4-Flash & OpenClaw   & 0.708 & 0.732 & 64 \\
Claude Opus 4.8   & Codex      & 0.698 & 0.790 & 51 \\
\bottomrule
\end{tabular}
\end{table}

\end{document}